\documentclass{article}
     \PassOptionsToPackage{numbers, compress}{natbib}
     \usepackage[preprint]{neurips_2020}

\usepackage{graphicx} %Loading the package
\graphicspath{{NeuRIPS2020/}}
\usepackage[utf8]{inputenc} % allow utf-8 input
\usepackage[T1]{fontenc}    % use 8-bit T1 fonts
\usepackage{hyperref}       % hyperlinks
\usepackage{url}            % simple URL typesetting
\usepackage{booktabs}       % professional-quality tables
\usepackage{amsfonts}       % blackboard math symbols
\usepackage{nicefrac}       % compact symbols for 1/2, etc.
\usepackage{microtype}      % microtypography
\usepackage{xcolor}
\usepackage{subcaption}
\usepackage[ruled,vlined]{algorithm2e}
\usepackage{mathrsfs}
\usepackage{multirow}
\usepackage[flushleft]{threeparttable}

\usepackage[colorinlistoftodos]{todonotes}

\title{Domain Generalization via Semi-supervised Meta Learning}

\author{%
  Hossein Sharifi-Noghabi, Hossein Asghari, Nazanin Mehrasa, Martin Ester\\
  School of Computing Science, Simon Fraser University\\
  Burnaby, British Columbia, Canada \\
  \texttt{[hsharifi,hasghari,nmehrasa,ester]@sfu.ca} \\
}

\begin{document}

\maketitle

\begin{abstract}
The goal of domain generalization is to learn from multiple source domains to generalize to unseen target domains under distribution discrepancy. Current state-of-the-art methods in this area are fully supervised, but for many real-world problems it is hardly possible to obtain enough labeled samples.
In this paper, we propose the first method of domain generalization to leverage unlabeled samples, combining of meta learning's episodic training and semi-supervised learning, called DGSML. DGSML employs an entropy-based pseudo-labeling approach to assign labels to unlabeled samples and then utilizes a novel discrepancy loss to ensure that class centroids before and after labeling unlabeled samples are close to each other. To learn a domain-invariant representation, it also utilizes a novel alignment loss to ensure that the distance between pairs of class centroids, computed after adding the unlabeled samples, is preserved across different domains. DGSML is trained by a meta learning approach to mimic the distribution shift between the input source domains and unseen target domains. Experimental results on benchmark datasets indicate that DGSML outperforms state-of-the-art domain generalization and semi-supervised learning methods.
\end{abstract}

\section{Introduction}
Deep neural networks have shown great performance in tasks with abundant labeled samples \citep{goodfellow2016deep,krizhevsky2012imagenet}. However, two major challenges exist in order to apply these networks to real-world tasks: first, different domains associated with a task have different distributions which violates the i.i.d assumption (train and test data are from the same distribution) and decreases the generalization capability of a model trained for that task \citep{rabanser2019failing}. Second, for most of the real-world tasks, labeling the data is either difficult or impossible resulting in a huge number of unlabeled samples. %For example, in drug response prediction, treating thousands of cancer patients with a drug might not be feasible or it is impossible to study the impact of a new drug on previous patients who have left the hospital, consequently, a majority of patient samples will be unlabeled. 
The question is can we employ both labeled and unlabeled samples from different domains, with different distributions, to train a model that generalizes to unseen domains? 

The answer to this question lies on the intersection of transfer learning \citep{pan2010survey} and semi-supervised learning \citep{sajjadi2016regularization,lee2013pseudo,ren2018meta,sohn2020fixmatch}. Transfer learning attempts to address domains' discrepancy by leveraging a domain-invariant representation across input domains, with different distributions, associated with the task of interest and semi-supervised learning attempts to leverage unlabeled data to boost the performance of a model on the task of interest. Semi-supervised learning approaches, such as consistency regularization \citep{sajjadi2016regularization} and pseudo-labeling \citep{lee2013pseudo}, do not consider distribution discrepancy and generalization to unseen target domains. On the other hand, while transfer learning approaches such as domain adaptation have shown great performance in addressing different domain discrepancies \citep{tzeng2014deep,chen2017no,tzeng2017adversarial,tsai2018learning,long2018conditional,pei2018multi,peng2018moment,peng2019domain,azizzadenesheli2019regularized,You_2019_CVPR}, they do not consider a combination of both labeled and unlabeled source domains. More importantly, the majority of transfer learning approaches assume that target domain is available during the training. Therefore, these areas cannot answer the above question separately.

The closest transfer learning approach to our goal is domain generalization \citep{matsuura2019domain,tseng2020cross,li2019episodic,dou2019domain,li2019feature,balaji2018metareg,ghifary2015domain,li2018domain,shankar2018generalizing,carlucci2019domain}. Domain generalization assumes that the target data is not available during training and the model should learn a domain-invariant representation only using the source domains with different distributions. Recent methods of domain generalization have adopted meta learning's episodic training to mimic domain shift by splitting the source domains into meta-train and meta-test at each iteration \citep{tseng2020cross,li2019episodic,dou2019domain,li2019feature,balaji2018metareg}. However, current state-of-the-art of domain generalization do not consider both labeled and unlabeled source domains, therefore, a need exists for a method that 1) employs both labeled and unlabeled samples, 2) generalizes to unseen target data, and 3) learns a domain-invariant predictive representation. 

In this paper, we propose DGSML, the first method of Domain Generalization based on Semi-supervised Meta Learning. To achieve a domain-invariant predictive representation, we propose a semi-supervised loss that combines entropy-based pseudo-labeling to assign labels to unlabeled samples and a discrepancy loss between class centroids (class means) with and without unlabeled samples. 
We also propose an alignment loss to minimize the discrepancy between the distance vector of the class centroids in one domain and the distance vector of the centroids in other domains. 
We demonstrate a significantly better accuracy compared to the state-of-the art methods of domain generalization and semi-supervised learning on two common benchmarks of domain generalization. 
%We evaluated the performance of DGSML on state-of-the-art domain generalization benchmarks for image classification and object recognition and observed a better performance compared to state-of-the art methods of domain generalization and semi-supervised learning.

\section{Related work}
\textbf{Domain adaptation} attempts to minimize the discrepancy between a labeled source domain \citep{tzeng2014deep,tzeng2017adversarial} (or multiple source domains \citep{peng2018moment}) and an unlabeled target \citep{chen2017no,pei2018multi} (or multiple targets \citep{peng2019domain}) domain and also minimize the prediction error on the labeled source domain as a proxy for the target domain. Common approaches to minimize the discrepancy are by utilizing discrepancy metrics such as the MMD \citep{tzeng2014deep,borgwardt2006integrating} or via adversarial learning \citep{chen2017no,tzeng2017adversarial,tsai2018learning,long2018conditional,pei2018multi,peng2019domain,azizzadenesheli2019regularized,You_2019_CVPR}. Based on the label space of source and target domain, domain adaptation can be closed set, partial \citep{cao2018partial}, open set \citep{panareda2017open}, or universal \citep{You_2019_CVPR}. %In the closed set domain adaptation, the label spaces of source and target domain are identical. In partial domain adaptation, source domain has private classes to target domain. In open set domain adaptation, both source and target domains can have private class to each other. Finally, in the universal domain adaptation there is no prior knowledge on the label space of the target domain. 

%\textbf{Multiple domain learning} attempts to learn a domain-invariant representation that minimizes the average prediction error for all of the input domains. Multiple domain learning is different from domain adaptation because methods of domain adaptation aim at learning a domain-invariant representation that minimizes the error on the target domain. Methods of multiple domain learning have a better chance of generalizing to unseen domains because they can leverage more labeled and unlabeled information and also they enable better knowledge transfer between the input domains. 

\textbf{Meta learning} attempts to learn how to train a model when a few labeled examples are available per class \citep{tseng2020cross,vinyals2016matching,finn2017model,chen2018a,scott2018adapted,snell2017prototypical}. An episode is a core idea of meta learning where each episode has a support set and a query set \citep{finn2017model}. The model is trained on the support set and then evaluated on the query set. Common approaches to meta learning are initialization-based methods and metric-based methods. In initialization-based methods, the idea is to provide a good initialization for the parameters such that the model generalizes to new classes with limited available samples as well as a few gradient steps. Model-agnostic
meta learning (MAML) \citep{finn2017model} is a well-known example of this category. In metric-based methods, the idea is to employ similarity metrics such as the Euclidean distance to guide the model to learn a representation that samples of the same class cluster closer to each other and far from those of the other classes. Prototypical Network (ProtoNet) \citep{snell2017prototypical} is a well-known example of this category. %that uses class centroids and the Euclidean distance to assign class labels. 
Although methods of meta learning have shown great performance in within domain generalization, the performance of these methods drops significantly under domain discrepancy \citep{chen2018a}. Moreover, current methods to address this discrepancy assume that either the target domain is accessible during meta-test or the input domains are entirely labeled \citep{tseng2020cross}.

\textbf{Semi-supervised learning} attempts to leverage unlabeled data during training. Common approaches to semi-supervised learning are consistency regularization \citep{sajjadi2016regularization} and pseudo-labeling \citep{lee2013pseudo}. In consistency regularization, the model predicts labels for the unlabeled samples and these predictions should be consistent for the perturbed version of the same samples. In pseudo-labeling, the idea is to utilize the predicted labels by the model for unlabeled samples with high confidence (e.g. above a certain threshold) and use those samples and their predicted pseudo-labels in retraining the model. 
A recent study showed that combining both consistency regularization and pseudo-labeling improves the state-of-the-art performance in semi-supervised learning benchmarks \citep{sohn2020fixmatch}. Moreover, incorporating pseudo-labeling in meta learning in semi-supervised ProtoNet has shown that utilizing both labeled and unlabeled data improves the performance of the models trained on only the labeled data \citep{ren2018meta}. this method assigns labels based on the Euclidean distance to the class centroids obtained from the labeled data. These centroids are then updated using the pseudo-labels assigned to the unlabeled data.

\textbf{Domain generalization} attempts to learn a domain-invariant representation given input data from multiple domains \citep{ghifary2015domain}. However, unlike domain adaptation, in domain generalization target domain is not available during training. This is a much harder scenario compared to domain adaptation where the target domain is available during the training \citep{dou2019domain}. A domain generalization method should extract a domain-invariant representation only using source domains. Domain generalization is important because it has similar settings as most of the real-world tasks for which no information is available about unseen data
\citep{shao2019regularized,matsuura2019domain,tseng2020cross,li2019episodic,dou2019domain,li2019feature,balaji2018metareg,ghifary2015domain,li2018domain,shankar2018generalizing,carlucci2019domain}. Domain generalization can be categorized into homogeneous and heterogeneous. In the homogeneous category, a shared label space exists between source domains and unseen target domains \citep{li2019episodic,dou2019domain,balaji2018metareg,ghifary2015domain,li2018domain,shankar2018generalizing,carlucci2019domain}, however, in the heterogeneous category the label spaces are disjoint \citep{tseng2020cross,li2019feature}.  
For example, Dou et al. \citep{dou2019domain} proposed model-agnostic learning of semantic features (MASF), a method based on meta learning to perform global and local alignment between domains in the homogeneous setting. This method uses class-specific mean and a Kullback–Leibler (KL) divergence in the global alignment step and a triplet loss for the local alignment between input domains. The role of meta learning is to utilize episodic training to generalize better under domain shift. %Tseng et al. \citep{tseng2020cross} proposed a model agnostic feature-wise transformation layer to enforce learning more diverse features and to avoid over-fitting to the input domains in the heterogeneous setting. They employed a meta learning approach to optimize the hyper-parameters of the feature-wise transformation layer and showed that such layers can be incorporated to the feature extractor of state-of-the-art meta learning methods and generalize better to unseen domains with discrepancy. 
Domain generalization is the closest related work to the goals of this research, nonetheless, a need still exists for a novel method that takes both labeled and unlabeled samples from different domains and learns a domain-invariant predictive representation. 

\section{Method}
\subsection{Problem definition}
Given $N$ source domains $D=\{D_{1},D_{2},...,D_{N}\}$ from different distributions on a joint space $\mathbb{X} \times \mathbb{Y}$, where $\mathbb{X}$ is an input space and $\mathbb{Y}$ is a label space, domain generalization assumes that a domain-invariant predictive feature space $\mathbb{Z} \in \mathbb{R^{D}}$ exists that generalizes to seen and unseen domains. 

We can use $D$ to train a feature extractor $F_{\theta}:\mathbb{X} \rightarrow \mathbb{Z}$ parameterized by $\theta$ that maps the input to the feature space and a classification task $T_{\phi}:\mathbb{Z} \rightarrow \mathbb{R^{C}}$ parameterized by $\phi$ that maps the extracted features to $C$ possible class labels. This can be achieved by optimizing $(\theta,\phi)$ via a task loss such as cross-entropy which will lead to a predictive $\mathbb{Z}$, however, such a representation cannot generalize to unseen target domains and will over-fit to labeled samples in $D$ without exploiting unlabeled samples.  

In the presence of unlabeled samples, we consider a partially unlabeled scenario. In a partially unlabeled scenario, each domain $D_{n}=\left(L_{l=1}^{M_{L}} \cup U_{u=1}^{M_{U}}\right)$, where $L$ consists of $M_{L}$ labeled samples $(x_{l}^{n},y_{l}^{n})$ and $U$ consists of $M_{U}$ unlabeled samples $(x_{u}^{n})$. The goal of DGSML is to learn a predictive representation by exploiting both labeled and unlabeled samples in $D$ with generalization capability to unseen target domains. 

\subsection{DGSML: Domain Generalization via Semi-supervised Meta Learning}
DGSML takes labeled and unlabeled samples from multiple source domains, learns a domain-invariant predictive representation with generalization capability to unseen target domains via a novel meta semi-supervised learning approach. DGSML achieves this by adopting an episodic training approach (meta learning) and three loss functions including, a task-specific loss for standard supervised classification, a semi-supervised loss based on a discrepancy loss between class centroids without unlabeled samples and with unlabeled samples, and an alignment loss that makes the distance vector of a class centroid to the other centroids in one domain to be similar to the distance vector of the same centroid in the other domains. These class centroids are obtained by considering both labeled samples and unlabeled samples assigned to the classes based on the semi-supervised loss. 

The meta learning approach splits the input source domains into meta-train, denoted by $D_{tr}$, and meta-test, denoted by $D_{ts}$, to mimic the distribution shift that we will face in unseen target domains and the loss functions are responsible for learning the domain-invariant predictive representation with generalization capability by employing both labeled and unlabeled samples. The semi-supervised loss is employed in the meta-train phase, the alignment loss function is employed in the meta-test phase, and the task-specific loss is employed in both phases. The task-specific loss and the semi-supervised loss ensure learning a predictive representation using labeled and unlabeled samples and the alignment loss function ensures that this representation is domain-invariant and generalizable to unseen target domains. The meta-train and meta-test phases are related to intra- and inter-domain updates and is different from standard meta learning methods like MAML. The most important connection to meta learning is the episodic training to resemble the distribution shift that the model will face in the deployment time on the target. Figure 1 and Algorithm 1 present the overview of DGSML.  

\begin{figure}
  \centering
  \includegraphics[width=\linewidth]{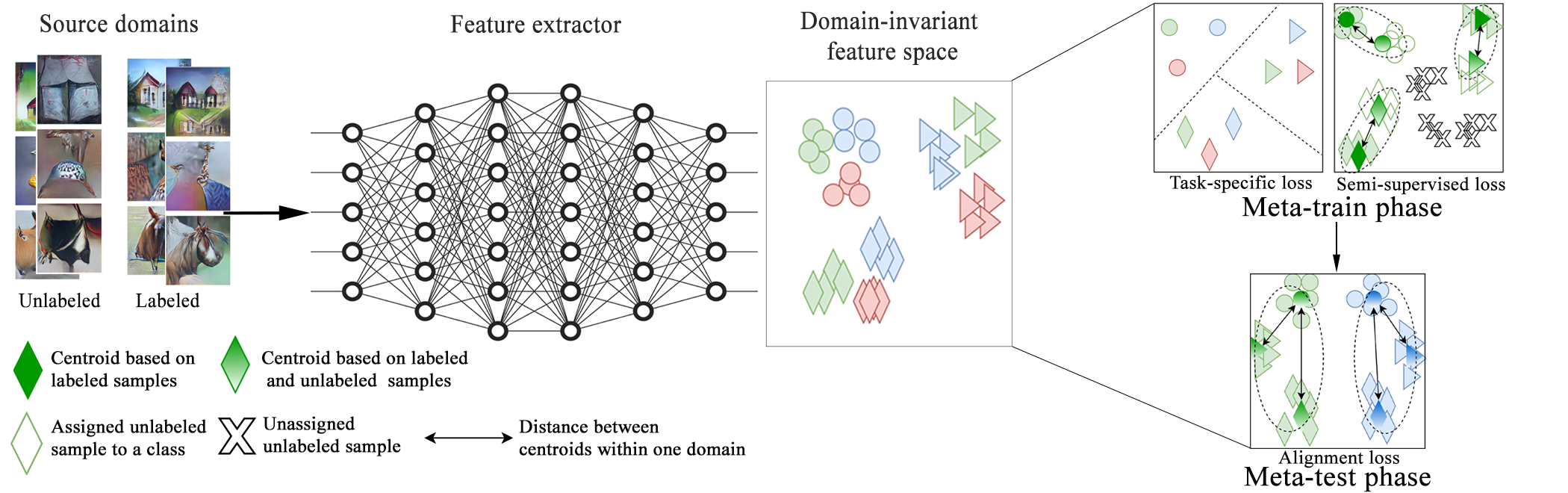}
  \caption{Schematic overview of DGSML; each color indicates a source domain in the feature space and each shape represents a class. The task-specific loss makes prediction based on labeled samples. The semi-supervised loss exploits unlabeled samples by minimizing the discrepancy between class centroids without unlabeled samples and with unlabeled samples. The alignment loss minimizes the difference of the pair-wise distances of class centroids in one domain to other domains.}
\end{figure}

\subsection{Meta-train phase}
The goal of meta-train phase is to learn a predictive representation with the use of both labeled and unlabeled samples in the source domains. This objective is achieved by employing a task-specific loss and a semi-supervised loss. 
\subsubsection{Task-specific loss}
The goal of this loss is to first map the input source domains to a lower-dimensional representation that is predictive of the class labels. To extract features, we designed a feature extractor as follows: 
%, \sigma(a) = e^{(a)} / \sum_{j} e^{(a_{j})}
\begin{equation}
    \overline{y_{i}}=\sigma(T_{\phi}(F_{\theta}(x_{i})), x_{i} \in D_{tr}=\{D_{1},D_{2},...,D_{|tr|}\}, 
\end{equation}
where, $F_{\theta}(x)$ denotes the feature extractor on $x_{i} \in D_{tr_{i}}$, $\sigma$ is the softmax activation over the output of a task-specific module denoted by $T_{\phi}$. With the final class predictions $\overline{y_{i}}$ available, we defined the task-specific loss for a labeled sample $(x_{i},y_{i}) \in D_{tr}$ from class $c$ using the standard cross-entropy as follows: 
\begin{equation}
    l_{task}(y,\overline{y};\theta,\phi)=-\sum\nolimits_{c} \textbf{1}[y=c]\log \overline{y}_{c}
\end{equation}

\subsubsection{Semi-supervised loss}
The goal of this loss is to exploit unlabeled samples in learning a predictive representation. We designed a loss function that employs discrepancy between class centroids without unlabeled samples and with unlabeled samples and also an entropy-based pseudo-labeling to enhance this. For pseudo-labeling, the idea is that an unlabeled sample such as $x_{u}$ is being labeled by utilizing $T_{\phi}$ as follows:
\begin{equation}
    \overline{y}_{u}=\sigma(T_{\phi}(F_{\theta}(x_{u}))), \sigma(a) = e^{(a)} / \sum\nolimits_{j} e^{(a_{j})}
\end{equation}
%\begin{equation}
 %   \overline{y}_{u}=\frac{e^{(-d(F_{\theta}(x_{u}),\mathscr{C}_{l}^{c}))}}{\sum_{c\prime}e^{(-d(F_{\theta}(x_{u}),\mathscr{C}%_{l}^{c\prime}))}},
%\end{equation}
%Using the obtained centroids, $x_{u}$ will get a label based on its distance to the class centroids, where, $d(.)$ is an arbitrary distance function. We used the Euclidean distance for DGSML. 
Equation 3 gives a vector of probabilities, denoted by $\overline{y}_{u}$, over the membership to the classes. If $\overline{y}_{u}$ follows a distribution close to a uniform distribution, $x_{u}$ is likely to be an outlier. To measure the confidence of the predictions for $x_{u}$, we defined a weight $w_{u}$ for unlabeled sample $x_{u}$ as follows:
\begin{equation}
    w_{u}=1-H(\overline{y}_{u}),
\end{equation}
where, $H(p)=\sum_{j} p_{j}\log p_{j}$ is the entropy.

With pseudo-labels available, we define the semi-supervised loss based on the discrepancy between the class centroids without unlabeled samples and the centroids with unlabeled samples. The idea is that the distance between extracted features of the class centroids obtained from the labeled samples of $D_{tr}$ and the class centroids obtained from both labeled and unlabeled samples of $D_{tr}$ should be minimum. The first step is to obtain the class centroids as follows:    
\begin{equation}
    \mathscr{C}_{l,d_{tr}}^{c}=\frac{1}{N_{c}} \sum\nolimits_{(x_{i},y_{i}=c)} F_{\theta}(x_{i}),
\end{equation}
where, $\mathscr{C}_{l,d_{tr}}^{c}$ denotes the class centroid for class $c$ based on the labeled samples of $d_{tr} \in D_{tr}$.\\
The class centroids based on both labeled and unlabeled samples can be obtained as follows: 
\begin{equation}
    \mathscr{C}_{l+u,d_{tr}}^{c}=\frac{\sum_{(x_{i},y_{i})\in L^{D_{tr}}}F_{\theta}(x_{i})+\sum_{(x_{i},\overline{y}_{i}=c) \in U^{D_{tr}}} w_{i} F_{\theta}(x_{i})}{N_{c}+N_{\overline{y}_{i}=c}}, 
\end{equation}
where, $\mathscr{C}_{l+u,d_{tr}}^{c}$ is the obtained class centroid for $c$ after considering both labeled and unlabeled samples with $\overline{y}_{i}=c$ in $d_{tr} \in D_{tr}$. Finally, for this loss we have:
\begin{equation}
    l_{sl}(\mathscr{C}_{l,d_{tr}},\mathscr{C}_{l+u,d_{tr}};\theta,\phi)= \sum\nolimits_{c} d(\mathscr{C}^{c}_{l,d_{tr}},\mathscr{C}^{c}_{l+u,d_{tr}}),
\end{equation}
where, $l_{sl}$ denotes the semi-supervised loss that has pseudo-labeling because of $\overline{y}$ and $w$, and the discrepancy loss because of penalizing the distance between the centroids. Finally, we defined the distance function $d(a,b)$ to be the $l^{2}$ norm of $(a-b)$. 

The parameters of $F_{\theta}(.)$ and $T_{\phi}(.)$ are optimized with gradient descent updates as follows:
\begin{equation}
    l_{meta-train} = l_{task}(y,\overline{y};\theta,\phi)+\beta_{0}l_{sl}(\mathscr{C}_{l,d_{tr}},\mathscr{C}_{l+u,d_{tr}};\theta,\phi)
\end{equation}
\begin{equation} 
    (\theta\prime,\phi\prime)= (\theta,\phi) - \alpha_{0}\nabla [l_{meta-train}],
\end{equation}
where, $\alpha_{0}$ is the learning rate and $\beta_{0}$ is the regularization coefficient. 
\subsection{Meta-test phase}
The goal of the meta-test phase is to learn a domain-invariant representation with generalization capability to unseen domains. We formulate alignment requirement for this representation as follows:\\
$d(\mathscr{C}_{l+u,d_{tr}}^{c_{1}},\mathscr{C}_{l+u,d_{tr}}^{c_{2}})\approx d(\mathscr{C}_{l+u,d_{ts}}^{c_{1}},\mathscr{C}_{l+u,d_{ts}}^{c_{2}})$, where $c_{1},c_{2} \in {1,...,C}$ and $d_{tr},d_{ts} \in D_{tr} \times D_{ts}$. \\
%2. Class-wise alignment: $\mathscr{C}_{l+u,d_{tr}}^{c}$ $\approx \mathscr{C}_{l+u,,d_{ts}}^{c}$, where $c \in {1,...,C}$ and $d_{tr},d_{ts} \in D_{tr} \times D_{ts}$. \\
Alignment means that the distance between two centroids in one domain should be similar to the distance between the same centroids in another domain. This condition makes the input domains to be similar in a global level.% we also need samples from similar classes across different domains to cluster together. %The class-wise alignment ensures samples of the same class but in different domains be close to each other in the feature space. This condition means that two centroids associated with the same class should be similar to each other, regardless of their corresponding domains. 
We designed an alignment loss function to address this condition. 
\subsubsection{Alignment loss}
The goal of this loss is to make the learning representation domain-invariant in the global level. This can be achieved by adversarial learning \citep{matsuura2019domain} or a soft confusion-matrix \citep{dou2019domain}. The confusion-matrix approach is based on the (similarity of the) distribution of predicted class labels meaning that a misclassified sample in $D_{i}$ should also be misclassified in $D_{j}$. However, similar prediction does not necessarily mean similar features. Therefore, in the representation level, we want distances between centroids in one domain $D_{i}$ to be similar to distances of the same centroids in another domain $D_{j}$. We formulate this requirement as follows:
\begin{equation}
    l_{alignment}(V_{\mathscr{C}};\theta\prime,\phi\prime) =\sum\nolimits_{c \in C} \sum\nolimits_{[d_{tr},d_{ts}] \in D_{tr} \times D_{ts}} d(V_{\mathscr{C}_{l+u,d_{tr}}^{c}},V_{\mathscr{C}_{l+u,d_{ts}}^{c}}),
\end{equation}
where $V_{\mathscr{C}^{c}}$ denotes the vector of pair-wise distance between the class centroid $\mathscr{C}^{c}$ in $ D_{i}$ and all the other centroids in $D_{i}$. This distance vector should be similar to the vector of $\mathscr{C}^{c}$ in $D_{j}$. Therefore, the alignment loss is the distance between the pair-wise distance vectors of class centroids in source domains. 
%\subsubsection{Class-wise alignment loss}
%The goal of this loss is to further regularize learning a domain-invariant representation in the class level. This means that samples of the same class should cluster together regardless of their domains. To achieve this, we design a loss function to minimize the distance between class centroids of the same class in different domains as follows:
%\begin{equation}
%    l_{class-wise}(\mathscr{C};\theta\prime,\phi\prime)=\sum_{c \in C} \sum_{[d_{tr}, d_{ts}] \in D_{tr} \times D_{ts}} d(\mathscr{C}_{l+u,d_{tr}}^{c}, \mathscr{C}_{l+u,d_{ts}}^{c}),
%\end{equation}
%similar to \citep{dou2019domain}, we employed a triplet loss function on the class centroids as follows: 
%\begin{equation}
  %  l_{class-wise}(A,N,P;\theta\prime,\phi\prime)=\sum_{A,N,P \in D_{tr} \times D_{ts}} %max[0,d(A,P)-d(A,N)+\lambda],
%\end{equation}
%where $A$ and $P$ are two class centroids of the same class in different domains and $N$ is a class centroid of another class and $\lambda$ is a margin. 
The total loss for the meta-test phase is:
\begin{equation}
    l_{meta-test} = l_{task}(d_{ts};\theta\prime,\phi\prime) + \beta_{1}l_{alignment} %+ %\beta_{2}l_{class-wise}.
\end{equation}
where, $l_{task}$ is obtained from labeled and unlabeled samples in $d_{ts}$, $\beta_{1}$ is the regularization coefficients. Finally, the parameters are optimized via incorporating both the meta-train and meta-test phases as follows: 
\begin{equation}
    (\theta,\phi) = (\theta,\phi) - \alpha_{1}\nabla [l_{meta-train}+l_{meta-test}] 
\end{equation}
\begin{algorithm}[t]
%\tiny
% \footnotesize
\SetAlgoLined
\textbf{Input:} $D=\{D_{1},D_{2},...,D_{N}\}$, $\alpha_{0}$, $\alpha_{1}$, $\beta_{0}$, $\beta_{1}$\\
\textbf{Output:} $F_{\theta}$, $T_{\phi}$ \\
 \While{not reached maximum iterations}{
  Randomly split D into $D_{tr}$ and $D_{ts}$ s.t. $D_{tr} \cap D_{ts}=\emptyset$ and  $D_{tr} \cup D_{ts}=D$\;
  \textbf{Meta-train on $D_{tr}$}\\
  \Indp
  Sample a mini-batch $d_{tr}$ from all domains in $D_{tr}$\;
  Calculate the supervised loss, $l_{task}(d_{tr};\theta,\phi)=-\sum_{c} \textbf{1}[y=c]\log \overline{y}_{c}$\;
  Calculate the semi-supervised loss, $l_{sl}(d_{tr};\theta,\phi)= \sum_{c} d(\mathscr{C}^{c}_{l,d_{tr}},\mathscr{C}^{c}_{l+u,d_{tr}})$\;
  calculate $l_{meta-train}=l_{task} + \beta_{0}l_{sl}$\;
  $(\theta\prime,\phi\prime) = (\theta,\phi) - \alpha_{0}\nabla [l_{meta-train}]$\;
  \Indm
  \textbf{Meta-test on $D_{ts}$}\\
  \Indp
  Sample a mini-batch $d_{ts}$ from all domains in $D_{ts}$\;
  Calculate $l_{task}(d_{ts},\theta\prime,\phi\prime)$\;
   Calculate $l_{alignment}$ using $(\theta\prime,\phi\prime)$ according to Eq. 11\;
   %Calculate $l_{class-wise}$ using $(\theta\prime,\phi\prime)$ according to Eq. 12\;
   Calculate $l_{meta-test} =l_{task} + \beta_{1}l_{alignment}$\; %+ \beta_{2}l_{class-wise}$\;
   $(\theta,\phi) = (\theta,\phi) - \alpha_{1}\nabla [l_{meta-train}+l_{meta-test}]$\;
   \Indm
 }
 \caption{DGSML}
\end{algorithm} 
\section{Experiments}
We designed our experiments to investigate whether incorporating unlabeled samples improves the prediction performance on unseen target domains as follows: 
1) We compared DGSML to  DeepAll as a simple but highly accurate baseline,  to semi-supervised ProtoNet (SSL-ProtoNet) \citep{ren2018meta} as a representative of state-of-the-art semi-supervised meta learning, to study domain generalization capability of our method, and to MASF \citep{dou2019domain}, as a representative of state-of-the-art fully labeled domain generalization, to study the impact of unlabeled samples in domain generalization capability of our method. 
2) We studied the impact of percentage of unlabeled samples in the performance of our method and the baselines. 
3) We performed an ablation study to investigate the contribution of each component of DGSML. 
We performed our experiments on the VLCS \citep{fang2013unbiased} domain generalization
benchmark for image classification and the PACS \citep{li2018domain} benchmark for object recognition. In our experiments on PACS and VLCS, we adopted a leave-one-domain-out scheme meaning that we considered three domains as the source domains and the fourth one as the unseen target domain. PACS and VLCS are the state-of-the-art benchmark datasets for domain generalization. More detail about the benchmarks is provided in the supplementary material. It is important to note that PACS and VLCS are fully labeled and we simulated unlabeled samples by withholding the class labels for different percentages of the samples. The unlabeled samples were selected randomly for each mini-batch/episode before the training. 
% \subsection{Baselines}

% SSL-ProtoNet \citep{ren2018meta} is a method of semi-supervised meta learning that assigns labels based on the Euclidean distance to the class centroids obtained from the labeled data. These centroids are then updated using the pseudo-labels assigned to the unlabeled data, and MASF \citep{dou2019domain} is a method of domain generalization based on meta learning and metric learning. 

For the partially unlabeled scenario, we discarded different percentages of labels in each source domain and treated them as unlabeled samples. Then, we used the remaining labeled samples to train MASF and DeepAll and employed the unlabeled and labeled samples to train SSL-ProtoNet and DGSML. It is important to note that MASF and DeepAll cannot incorporate unlabeled samples. Moreover, their performance in the fully labeled scenario should be considered as an upper bound for DGSML because unlike our method, they have access to all of the labels.  

We used AlexNet and ResNet-18 pre-trained on ImageNet for the feature extractor of DGSML and the baselines except for SSL-ProtoNet (We used four convolutional layers proposed by the original authors). We fine-tuned AlexNet but kept ResNet-18 frozen and did not fine-tune it. The last layer of both of them was modified to predict the same number of classes in each dataset via an additional classification layer. For DGSML, we adopted standard train/validation/test splits provided for PACS \citep{li2017deeper} and VLCS \citep{carlucci2019domain} and implemented it using the Pytorch framework. For each rate of unlabeled samples, we selected five random subsets with replacement (using five different seeds) and reported the average and standard error of the performance. The implementation details for the baselines and also the code and data to reproduce DGSML results are provided in the supplementary material.

\section{Results}
Tables 1 and 2 present the accuracy of DGSML compared to the studied baselines on VLCS and PACS datasets, respectively, using AlexNet. On VLCS, our method outperformed the baselines for most of the investigated unlabeled rates. We observed that the performance gap was larger in favor of DGSML for high rates of unlabeled samples ($95\%$) which indicates that DGSML makes generalizable predictions more accurately when it has access to more unlabeled samples. Moreover, DGSML had a lower standard error than the baselines which indicates it is more robust.  On PACS, MASF generally showed a better performance than DGSML, however, similar to VLCS, our method demonstrated a better performance for higher rates of unlabeled samples ($0.95\%$). In terms of robustness, our method showed a lower standard error on PACS dataset as well.We also compared our method to MASF and DeepAll when they were trained on all of the labeled samples available ($0\%$ rate of unlabeled samples, see Table S1 in the supplementary material). DGSML outperformed the baselines on VLCS dataset when it had access to $50\%$ of the labeled samples. We observed similar results for the $20\%$ scenario as well. 

DeepAll is known to be a surprisingly competitive baseline, possibly even better than some of the state-of-the-art methods of domain generalization \citep{li2019feature,li2017deeper}, which is confirmed by our strong results for this baseline. However, DeepAll requires many labeled samples to be more accurate. We observed that SSL-ProtoNet performed poorly in almost all of the experiments. We believe that this large performance gap, compared to the other methods, is due to the fact that SSL-ProtoNet has been designed for semi-supervised few-shot learning. The performance of methods of few-shot learning decreases significantly when we have a shallow feature extractor and also when our domains have distribution shifts \citep{chen2018a}. Therefore, the combination of a shallow backbone and domain shifts decreases the performance significantly. MASF and DGSML overall showed comparable results. We argue that this is due to the fact that MASF uses two levels of alignment (global and local), while DGSML employs only one level of alignment, but that DGSML exploits unlabeled samples, while MASF does not do so. Therefore, on the harder benchmark PACS, MASF was more accurate because the stronger alignment had more impact, and for VLCS with less domain discrepancy, unlabeled samples had more impact than stronger alignment, and consequently DGSML outperformed MASF. Finally, we performed an ablation study and confirmed that DGSML with all of its losses had the best performance compared to its variants (See Table S2 in the supplementary material).
\begin{table}
  %\tiny
  \caption{Accuracy on VLCS with different rates of unlabeled samples using shallow feature extractor}
  \label{vlcs-table}
  \centering
  \begin{tabular}{lllcccc}
    \toprule
    Rate & Source     & Target     & {MASF \citep{dou2019domain}} & {SSL-ProtoNet \citep{ren2018meta}} & {DeepAll} & {DGSML} \\
    \midrule
    \multirow{4}{*}{20\%} &  C,L,P & Sun  & {$64.78 \pm 0.58$} & {$26.55 \pm 0.57$}  & {$64.48 \pm 0.48$} & {$\mathbf{65.73 \pm 0.51}$}     \\
    & L,P,S & Caltech & {$93.41 \pm 0.29$} & {$45.14 \pm 2.84$} & {$94.96 \pm 0.17$} & {$\mathbf{95.49 \pm 0.40}$}       \\
    & C,P,S & Labelme & {$\mathbf{60.48 \pm 1.05}$} & {$24.23 \pm 0.99$} & {$55.38 \pm 0.63$} & {$58.90 \pm 0.66$}   \\
    & C,L,S & Pascal & {$\mathbf{68.66 \pm 0.50}$} & {$24.50 \pm 0.59$} & {$66.92 \pm 0.55$} & {$68.06 \pm 0.37$}   \\ 
    \midrule
    \multirow{4}{*}{50\%} & C,L,P & Sun  & {$64.07 \pm 0.55$} & {$29.46 \pm 1.25$}  & {$64.42 \pm 0.69$} & {$\mathbf{65.27 \pm 0.38}$}     \\
    & L,P,S & Caltech & {$92.03 \pm 0.49$} & {$44.95 \pm 2.51$} & {$94.50 \pm 0.41$} & {$\mathbf{95.93 \pm 0.33}$}       \\
    & C,P,S & Labelme & {$\mathbf{60.16 \pm 0.61}$} & {$22.85 \pm 0.30$} & {$56.01 \pm 0.80$} & {$58.43 \pm 0.61$}  \\
    & C,L,S & Pascal & {$\mathbf{67.80 \pm 0.40}$} & {$24.80 \pm 0.62$} & {$65.75 \pm 0.75$} & {$67.32 \pm 0.43$}   \\    
    \midrule
    \multirow{4}{*}{80\%} & C,L,P & Sun  & {$64.07 \pm 0.81$} & {$28.47 \pm 1.22$}  & {$63.57 \pm 1.30$} & {$\mathbf{64.75 \pm 0.46}$}    \\
    & L,P,S & Caltech & {$91.73 \pm 0.80$} & {$42.93 \pm 0.56$} & {$93.64 \pm 0.47$} & {$\mathbf{95.11 \pm 0.14}$}       \\
    & C,P,S & Labelme & {$\mathbf{58.71 \pm 1.33}$} & {$25.77 \pm 1.36$} & {$54.34 \pm 0.76$} & {$57.71 \pm 0.51$}  \\
    & C,L,S & Pascal & {$65.59 \pm 0.62$} & {$23.90 \pm 0.71$} & {$64.75 \pm 0.36$} & {$\mathbf{67.00 \pm 0.25}$}  \\   
    \midrule
    \multirow{4}{*}{95\%} & C,L,P & Sun  & {$55.95 \pm 1.88$} & {$30.12 \pm 1.40$}  & {$55.19 \pm 2.20$} & {$\mathbf{63.17 \pm 0.62}$}     \\
    & L,P,S & Caltech & {$86.11 \pm 2.49$} & {$53.09 \pm 1.37$} & {$\mathbf{90.63 \pm 1.59}$} & {$90.42 \pm 1.82$}       \\
    & C,P,S & Labelme & {$53.56 \pm 0.65$} & {$29.01 \pm 0.64$} & {$55.76 \pm 1.85$} & {$\mathbf{57.67 \pm 1.08}$}  \\
    & C,L,S & Pascal & {$60.12 \pm 1.51$} & {$25.63 \pm 2.01$} & {$56.92 \pm 1.85$} & {$\mathbf{62.55 \pm 0.65}$}   \\    
    \midrule
    \multicolumn{3}{c}{Average} & 69.20 & 31.34 & 68.50 & 70.84 \\
    \bottomrule
  \end{tabular}
\end{table}

\begin{table}
  %\tiny
  \caption{Accuracy on PACS with different rates of unlabeled samples using shallow feature extractor}
  \label{pacs-table}
  \centering
  \begin{tabular}{lllcccc}
    \toprule
    Rate & Source     & Target     & {MASF \citep{dou2019domain}} & {SSL-ProtoNet \citep{ren2018meta}} & {DeepAll} & {DGSML} \\
    \midrule
    \multirow{4}{*}{20\%} &  A,C,S & Photo  & {$\mathbf{88.57 \pm 0.39}$} & {$39.51 \pm 0.96$} & {$87.17 \pm 0.32$} & {$88.40 \pm 0.10$}    \\
    & C,S,P & Art & {$\mathbf{66.60 \pm 0.28}$} & {$24.57 \pm 0.47$} & {$62.03 \pm 0.38$} & {$61.68 \pm 0.39$}    \\
    & A,C,P & Sketch & {$\mathbf{59.56 \pm 1.61}$} & {$43.43 \pm 1.10$} & {$55.67 \pm 1.43$} & {$50.01 \pm 0.84$} \\
    & A,S,P & Cartoon & {$\mathbf{69.34 \pm 0.87}$} & {$33.44 \pm 1.30$} & {$62.50 \pm 1.29$} & {$65.49 \pm 0.48$} \\
    \midrule
    \multirow{4}{*}{50\%} & A,C,S & Photo & {$86.80 \pm 0.62$} & {$37.33 \pm 1.62$}  & {$86.01 \pm 0.31$} & {$\mathbf{87.72 \pm 0.27}$}  \\
    & C,S,P & Art & {$\mathbf{64.29 \pm 0.82}$} & {$24.77 \pm 0.24$} & {$59.14 \pm 0.50$} & {$62.52 \pm 0.40$}    \\
    & A,C,P & Sketch & {$\mathbf{58.67 \pm 3.40}$} & {$41.74 \pm 1.69$} & {$52.56 \pm 3.40$} & {$50.38 \pm 1.25$}  \\
    & A,S,P & Cartoon & {$66.53 \pm 1.06$} & {$35.03 \pm 1.20$} & {$63.99 \pm 0.50$} & {$\mathbf{66.84 \pm 0.64}$}  \\    
    \midrule
    \multirow{4}{*}{80\%} & A,C,S & Photo  & {$\mathbf{83.28 \pm 1.37}$} & {$38.38 \pm 1.24$}  & {$82.53 \pm 1.27$} & {$83.19 \pm 0.23$}     \\
    & C,S,P & Art & {$61.26 \pm 0.59$} & {$24.81 \pm 0.48$} & {$56.04 \pm 0.55$} & {$59.22 \pm 0.64$} \\
    & A,C,P & Sketch & {$\mathbf{57.03 \pm 1.36}$} & {$39.16 \pm 1.68$} & {$48.42 \pm 3.34$} & {$46.37 \pm 2.50$}  \\
    & A,S,P & Cartoon & {$\mathbf{65.01 \pm 1.63}$} & {$33.98 \pm 1.28$} & {$60.04 \pm 1.21$} & {$64.81 \pm 0.75$}   \\    
    \midrule
    \multirow{4}{*}{95\%} & A,C,S & Photo  & {$75.32 \pm 1.12$} & {$36.89 \pm 1.90$}   & {$76.43 \pm 1.40$} & {$\mathbf{79.45 \pm 0.85}$} \\
    & C,S,P & Art & {$53.41 \pm 1.70$} & {$23.24 \pm 0.79$} & {$49.38 \pm 1.76$} & {$\mathbf{54.39 \pm 1.22}$} \\
    & A,C,P & Sketch & {$\mathbf{49.02 \pm 1.24}$} & {$42.21 \pm 1.92$}  & {$40.11 \pm 3.74$} & {$48.05 \pm 1.35$}   \\
    & A,S,P & Cartoon & {$61.37 \pm 1.34$} & {$34.58 \pm 1.09$} & {$58.83 \pm 1.28$} & {$\mathbf{62.54 \pm 0.67}$}   \\    
    \midrule
    \multicolumn{3}{c}{Average} & {66.63} & {34.57} & {62.50} & {64.44} \\
    \bottomrule
  \end{tabular}
\end{table}
\subsection{Deeper feature extractor}
To study the performance of DGSML and MASF with a deeper feature extractor, we tested ResNet-18 on the PACS benchmark (see supplementary material for implementation detail) and present the results in Table 3. We observed noticeable improvements in the performance of both DGSML and MASF compared to the results with AlexNet. Using this deeper architecture, DGSML was able to consistently and significantly outperform MASF for all rates of unlabeled samples. This suggests that for datasets with high domain discrepancies such as PACS, the combination of leveraging unlabeled samples and more abstract features, obtained via a deeper architecture, plays a beneficial role in domain generalization. As expected, we also observed that the performance of SSL-ProtoNet on PACS increased significantly from $34\%$ using the shallow backbone to $62\%$ using ResNet (See Table S3 in the supplementary material for detailed results). 

\begin{table}
  \centering
  \caption{Accuracy on PACS with different rates of unlabeled samples using ResNet-18}
  \begin{threeparttable}
  \label{resnet-table}
  %\tiny
  \begin{tabular}{ll|lcc|lcc}
    \toprule
    Source     & Target & Rate & {MASF \citep{dou2019domain}} & {DGSML} & Rate & {MASF \citep{dou2019domain}} & {DGSML} \\
    \midrule
    A,C,S & Photo & \multirow{4}{*}{20\%} &  {$94.12 \pm 0.11$} & {$\mathbf{95.14 \pm 0.14}$} & \multirow{4}{*}{80\%} & {$93.99 \pm 0.16$} & {$\mathbf{95.01 \pm 0.10}$}    \\
    C,S,P & Art & & {$68.35 \pm 0.39$} & {$\mathbf{72.57 \pm 0.10}$} & & {$67.91 \pm 0.28$} & {$\mathbf{72.03 \pm 0.24}$}    \\
    A,C,P & Sketch & & {$49.83 \pm 0.39$} & {$\mathbf{53.09 \pm 0.34}$} & & {$49.32 \pm 0.70$} & {$\mathbf{52.37 \pm 0.91}$} \\
    A,S,P & Cartoon & & {$70.02 \pm 0.53$} & {$\mathbf{73.62 \pm 0.13}$} & & {$69.77 \pm 0.45$} & {$\mathbf{73.30 \pm 0.32}$} \\
    \midrule
    A,C,S & Photo & \multirow{4}{*}{50\%} &  {$94.40 \pm 0.18$} & {$\mathbf{95.08 \pm 0.07}$}  & \multirow{4}{*}{95\%} &  {$92.87 \pm 0.39$} & {$\mathbf{94.12 \pm 0.21}$}  \\
    C,S,P & Art & & {$68.34 \pm 0.26$} & {$\mathbf{72.54 \pm 0.24}$} & & {$66.47 \pm 0.67$} & {$\mathbf{70.55 \pm 0.20}$}    \\
    A,C,P & Sketch & & {$49.75 \pm 0.32$} & {$\mathbf{52.41 \pm 0.53}$} & & {$47.45 \pm 0.46$} & {$\mathbf{49.69 \pm 0.51}$}  \\
    A,S,P & Cartoon & & {$69.46 \pm 0.39$} & {$\mathbf{73.52 \pm 0.10}$} & & {$69.31 \pm 0.99$} & {$\mathbf{72.42 \pm 0.83}$}  \\    
    \midrule
    \multicolumn{2}{c}{Total average} & & MASF: & {70.09} & & DGSML: & {72.98}  \\
    \bottomrule
  \end{tabular}
  \begin{tablenotes}
      \small
      \item \tiny Note: Total average is calculated over all rates of unlabeled samples for each method. 
    \end{tablenotes}
  \end{threeparttable}
  \end{table}

\subsection{Discussion}
We chose AlexNet and ResNet-18 as feature extractors because they train faster compared to deeper networks and also because other methods of domain generalization utilized them frequently \citep{matsuura2019domain,dou2019domain,li2019feature,shao2019regularized}, however, DGSML can work with deeper architectures such as ResNet-50 as well. It is important to note that unlike AlexNet, we did not fine-tune ResNet-18 and kept it frozen to save computational resources. We expect to see additional improvements in DGSML and MASF performance by fine-tuning ResNet-18. Although DGSML employs second order gradients which might be slow, we were able to train it from scratch on a GeForce GTX 1080 GPU in under four hours for one AlexNet setting. %We did not perform a comprehensive search for hyper-parameters of DGSML, we believe DGSML performance will be improved by tuning hyper-parameters. 
In DGSML episodic training, samples are selected randomly from the pool of available labeled/unlabeled samples, therefore, there is no guarantee that the method has seen all of the samples during the training.  

DGSML obtained the best accuracy compared to the baselines when it had access to a lot of unlabeled samples, This observation indicates that if employing a shallow feature extractor like AlexNet, the benefit of the proposed method is for real-world applications when the number of labeled samples is much smaller than the number of unlabeled samples, and if employing a deeper feature extractor like ResNet-18, DGSML works accurately for diverse rates of unlabeled samples. One example of such a real-world application is drug response prediction where we have access to a lot of genomic data obtained from cancer patients but only a small portion of them are treated with cancer drugs and current state-of-the-art methods of this application cannot incorporate unlabeled samples \citep{hshn2020AITL}. In our future work, our goal is to apply DGSML to this real-world application.

\section{Conclusion}
In this paper, we presented DGSML, the first method of domain generalization based on episodic training in meta learning and semi-supervised learning. DGSML utilizes three loss functions: 1) the task-specific loss to make predictions, 2) the semi-supervised loss to ensure that class centroids before and after labeling unlabeled samples are similar, and 3) the alignment loss to ensure that the distance between the class centroids (after adding unlabeled samples) in one domain is similar to the distance of class centroids in other domains. The combination of these losses makes the representation predictive and domain-invariant by exploiting both labeled and unlabeled samples. We compared DGSML to state-of-the-art methods of fully supervised domain generalization and of semi-supervised meta learning and obtained promising results. Particularly, DGSML demonstrated the best results for any rates of unlabeled samples when it employed a deep feature extractor (ResNet-18) and when it utilized a shallow feature extractor (AlexNet) with the number of labeled samples much smaller than number of unlabeled samples.

\section*{Reproducibility}
Supplementary material, all the utilized datasets, code, and conda environments to re-run and reproduce our results are provided in the following Github repository:\\
 \url{https://github.com/hosseinshn/DGSML} 

\section*{Acknowledgement}
We would like to thank Oliver Snow, Shuman Peng, and Saeed Izadi (Simon Fraser University), Drs. Colin Collins and Art Cherkasov (the Vancouver Prostate Centre) and Dr. Bastian Rieck (ETH Zurich) for their support. We also would like to thank the Vancouver Prostate Centre and Compute Canada for providing the computational resources for this research. This research was supported by a Discovery Grant from the National Science and Engineering Research Council of Canada (to M.E.).

\section*{Authors' contributions}
Study concept and design: H.SN. and M.E.\\
DGSML development: H.SN.\\
Implementations: H.SN., H.A., and N.M.\\
Analysis and interpretation of results: H.SN., H.A., and N.M.\\
High performance computing: H.A.\\
Drafting of the manuscript: All authors read and approved the final manuscript.\\
Supervision: M.E.

\bibliographystyle{unsrtnat}
\bibliography{Ref}

\begin{thebibliography}{40}
\providecommand{\natexlab}[1]{#1}
\providecommand{\url}[1]{\texttt{#1}}
\expandafter\ifx\csname urlstyle\endcsname\relax
  \providecommand{\doi}[1]{doi: #1}\else
  \providecommand{\doi}{doi: \begingroup \urlstyle{rm}\Url}\fi

\bibitem[Goodfellow et~al.(2016)Goodfellow, Bengio, Courville, and
  Bengio]{goodfellow2016deep}
Ian Goodfellow, Yoshua Bengio, Aaron Courville, and Yoshua Bengio.
\newblock \emph{Deep learning}, volume~1.
\newblock MIT Press, 2016.

\bibitem[Krizhevsky et~al.(2012)Krizhevsky, Sutskever, and
  Hinton]{krizhevsky2012imagenet}
Alex Krizhevsky, Ilya Sutskever, and Geoffrey~E Hinton.
\newblock Imagenet classification with deep convolutional neural networks.
\newblock In \emph{Advances in neural information processing systems}, pages
  1097--1105, 2012.

\bibitem[Rabanser et~al.(2019)Rabanser, G{\"u}nnemann, and
  Lipton]{rabanser2019failing}
Stephan Rabanser, Stephan G{\"u}nnemann, and Zachary Lipton.
\newblock Failing loudly: an empirical study of methods for detecting dataset
  shift.
\newblock In \emph{Advances in Neural Information Processing Systems}, pages
  1394--1406, 2019.

\bibitem[Pan and Yang(2010)]{pan2010survey}
Sinno~Jialin Pan and Qiang Yang.
\newblock A survey on transfer learning.
\newblock \emph{IEEE Transactions on knowledge and data engineering},
  22\penalty0 (10):\penalty0 1345--1359, 2010.

\bibitem[Sajjadi et~al.(2016)Sajjadi, Javanmardi, and
  Tasdizen]{sajjadi2016regularization}
Mehdi Sajjadi, Mehran Javanmardi, and Tolga Tasdizen.
\newblock Regularization with stochastic transformations and perturbations for
  deep semi-supervised learning.
\newblock In \emph{Advances in neural information processing systems}, pages
  1163--1171, 2016.

\bibitem[Lee(2013)]{lee2013pseudo}
Dong-Hyun Lee.
\newblock Pseudo-label: The simple and efficient semi-supervised learning
  method for deep neural networks.
\newblock In \emph{Workshop on challenges in representation learning, ICML},
  volume~3, page~2, 2013.

\bibitem[Ren et~al.(2018)Ren, Triantafillou, Ravi, Snell, Swersky, Tenenbaum,
  Larochelle, and Zemel]{ren2018meta}
Mengye Ren, Eleni Triantafillou, Sachin Ravi, Jake Snell, Kevin Swersky,
  Joshua~B Tenenbaum, Hugo Larochelle, and Richard~S Zemel.
\newblock Meta-learning for semi-supervised few-shot classification.
\newblock \emph{arXiv preprint arXiv:1803.00676}, 2018.

\bibitem[Sohn et~al.(2020)Sohn, Berthelot, Li, Zhang, Carlini, Cubuk, Kurakin,
  Zhang, and Raffel]{sohn2020fixmatch}
Kihyuk Sohn, David Berthelot, Chun-Liang Li, Zizhao Zhang, Nicholas Carlini,
  Ekin~D Cubuk, Alex Kurakin, Han Zhang, and Colin Raffel.
\newblock Fixmatch: Simplifying semi-supervised learning with consistency and
  confidence.
\newblock \emph{arXiv preprint arXiv:2001.07685}, 2020.

\bibitem[Tzeng et~al.(2014)Tzeng, Hoffman, Zhang, Saenko, and
  Darrell]{tzeng2014deep}
Eric Tzeng, Judy Hoffman, Ning Zhang, Kate Saenko, and Trevor Darrell.
\newblock Deep domain confusion: Maximizing for domain invariance.
\newblock \emph{ICCV}, 2014.

\bibitem[Chen et~al.(2017)Chen, Chen, Chen, Tsai, Frank~Wang, and
  Sun]{chen2017no}
Yi-Hsin Chen, Wei-Yu Chen, Yu-Ting Chen, Bo-Cheng Tsai, Yu-Chiang Frank~Wang,
  and Min Sun.
\newblock No more discrimination: Cross city adaptation of road scene
  segmenters.
\newblock In \emph{Proceedings of the IEEE International Conference on Computer
  Vision}, pages 1992--2001, 2017.

\bibitem[Tzeng et~al.(2017)Tzeng, Hoffman, Saenko, and
  Darrell]{tzeng2017adversarial}
Eric Tzeng, Judy Hoffman, Kate Saenko, and Trevor Darrell.
\newblock Adversarial discriminative domain adaptation.
\newblock In \emph{Proceedings of the IEEE Conference on Computer Vision and
  Pattern Recognition}, pages 7167--7176, 2017.

\bibitem[Tsai et~al.(2018)Tsai, Hung, Schulter, Sohn, Yang, and
  Chandraker]{tsai2018learning}
Yi-Hsuan Tsai, Wei-Chih Hung, Samuel Schulter, Kihyuk Sohn, Ming-Hsuan Yang,
  and Manmohan Chandraker.
\newblock Learning to adapt structured output space for semantic segmentation.
\newblock In \emph{Proceedings of the IEEE Conference on Computer Vision and
  Pattern Recognition}, pages 7472--7481, 2018.

\bibitem[Long et~al.(2018)Long, Cao, Wang, and Jordan]{long2018conditional}
Mingsheng Long, Zhangjie Cao, Jianmin Wang, and Michael~I Jordan.
\newblock Conditional adversarial domain adaptation.
\newblock In \emph{Advances in Neural Information Processing Systems}, pages
  1640--1650, 2018.

\bibitem[Pei et~al.(2018)Pei, Cao, Long, and Wang]{pei2018multi}
Zhongyi Pei, Zhangjie Cao, Mingsheng Long, and Jianmin Wang.
\newblock Multi-adversarial domain adaptation.
\newblock In \emph{Thirty-Second AAAI Conference on Artificial Intelligence},
  2018.

\bibitem[Peng et~al.(2018)Peng, Bai, Xia, Huang, Saenko, and
  Wang]{peng2018moment}
Xingchao Peng, Qinxun Bai, Xide Xia, Zijun Huang, Kate Saenko, and Bo~Wang.
\newblock Moment matching for multi-source domain adaptation.
\newblock \emph{arXiv preprint arXiv:1812.01754}, 2018.

\bibitem[Peng et~al.(2019)Peng, Huang, Sun, and Saenko]{peng2019domain}
Xingchao Peng, Zijun Huang, Ximeng Sun, and Kate Saenko.
\newblock Domain agnostic learning with disentangled representations.
\newblock \emph{ICML}, 2019.

\bibitem[Azizzadenesheli et~al.(2019)Azizzadenesheli, Liu, Yang, and
  Anandkumar]{azizzadenesheli2019regularized}
Kamyar Azizzadenesheli, Anqi Liu, Fanny Yang, and Animashree Anandkumar.
\newblock Regularized learning for domain adaptation under label shifts.
\newblock \emph{arXiv preprint arXiv:1903.09734}, 2019.

\bibitem[You et~al.(2019)You, Long, Cao, Wang, and Jordan]{You_2019_CVPR}
Kaichao You, Mingsheng Long, Zhangjie Cao, Jianmin Wang, and Michael~I. Jordan.
\newblock Universal domain adaptation.
\newblock In \emph{The IEEE Conference on Computer Vision and Pattern
  Recognition (CVPR)}, June 2019.

\bibitem[Matsuura and Harada(2019)]{matsuura2019domain}
Toshihiko Matsuura and Tatsuya Harada.
\newblock Domain generalization using a mixture of multiple latent domains.
\newblock \emph{arXiv preprint arXiv:1911.07661}, 2019.

\bibitem[Tseng et~al.(2020)Tseng, Lee, Huang, and Yang]{tseng2020cross}
Hung-Yu Tseng, Hsin-Ying Lee, Jia-Bin Huang, and Ming-Hsuan Yang.
\newblock Cross-domain few-shot classification via learned feature-wise
  transformation.
\newblock \emph{arXiv preprint arXiv:2001.08735}, 2020.

\bibitem[Li et~al.(2019{\natexlab{a}})Li, Zhang, Yang, Liu, Song, and
  Hospedales]{li2019episodic}
Da~Li, Jianshu Zhang, Yongxin Yang, Cong Liu, Yi-Zhe Song, and Timothy~M
  Hospedales.
\newblock Episodic training for domain generalization.
\newblock In \emph{Proceedings of the IEEE International Conference on Computer
  Vision}, pages 1446--1455, 2019{\natexlab{a}}.

\bibitem[Dou et~al.(2019)Dou, de~Castro, Kamnitsas, and Glocker]{dou2019domain}
Qi~Dou, Daniel~Coelho de~Castro, Konstantinos Kamnitsas, and Ben Glocker.
\newblock Domain generalization via model-agnostic learning of semantic
  features.
\newblock In \emph{Advances in Neural Information Processing Systems}, pages
  6447--6458, 2019.

\bibitem[Li et~al.(2019{\natexlab{b}})Li, Yang, Zhou, and
  Hospedales]{li2019feature}
Yiying Li, Yongxin Yang, Wei Zhou, and Timothy~M Hospedales.
\newblock Feature-critic networks for heterogeneous domain generalization.
\newblock \emph{arXiv preprint arXiv:1901.11448}, 2019{\natexlab{b}}.

\bibitem[Balaji et~al.(2018)Balaji, Sankaranarayanan, and
  Chellappa]{balaji2018metareg}
Yogesh Balaji, Swami Sankaranarayanan, and Rama Chellappa.
\newblock Metareg: Towards domain generalization using meta-regularization.
\newblock In \emph{Advances in Neural Information Processing Systems}, pages
  998--1008, 2018.

\bibitem[Ghifary et~al.(2015)Ghifary, Bastiaan~Kleijn, Zhang, and
  Balduzzi]{ghifary2015domain}
Muhammad Ghifary, W~Bastiaan~Kleijn, Mengjie Zhang, and David Balduzzi.
\newblock Domain generalization for object recognition with multi-task
  autoencoders.
\newblock In \emph{Proceedings of the IEEE international conference on computer
  vision}, pages 2551--2559, 2015.

\bibitem[Li et~al.(2018)Li, Jialin~Pan, Wang, and Kot]{li2018domain}
Haoliang Li, Sinno Jialin~Pan, Shiqi Wang, and Alex~C Kot.
\newblock Domain generalization with adversarial feature learning.
\newblock In \emph{Proceedings of the IEEE Conference on Computer Vision and
  Pattern Recognition}, pages 5400--5409, 2018.

\bibitem[Shankar et~al.(2018)Shankar, Piratla, Chakrabarti, Chaudhuri, Jyothi,
  and Sarawagi]{shankar2018generalizing}
Shiv Shankar, Vihari Piratla, Soumen Chakrabarti, Siddhartha Chaudhuri, Preethi
  Jyothi, and Sunita Sarawagi.
\newblock Generalizing across domains via cross-gradient training.
\newblock \emph{arXiv preprint arXiv:1804.10745}, 2018.

\bibitem[Carlucci et~al.(2019)Carlucci, D'Innocente, Bucci, Caputo, and
  Tommasi]{carlucci2019domain}
Fabio~M Carlucci, Antonio D'Innocente, Silvia Bucci, Barbara Caputo, and
  Tatiana Tommasi.
\newblock Domain generalization by solving jigsaw puzzles.
\newblock In \emph{Proceedings of the IEEE Conference on Computer Vision and
  Pattern Recognition}, pages 2229--2238, 2019.

\bibitem[Borgwardt et~al.(2006)Borgwardt, Gretton, Rasch, Kriegel,
  Sch{\"o}lkopf, and Smola]{borgwardt2006integrating}
Karsten~M Borgwardt, Arthur Gretton, Malte~J Rasch, Hans-Peter Kriegel,
  Bernhard Sch{\"o}lkopf, and Alex~J Smola.
\newblock Integrating structured biological data by kernel maximum mean
  discrepancy.
\newblock \emph{Bioinformatics}, 22\penalty0 (14):\penalty0 e49--e57, 2006.

\bibitem[Cao et~al.(2018)Cao, Long, Wang, and Jordan]{cao2018partial}
Zhangjie Cao, Mingsheng Long, Jianmin Wang, and Michael~I Jordan.
\newblock Partial transfer learning with selective adversarial networks.
\newblock In \emph{Proceedings of the IEEE Conference on Computer Vision and
  Pattern Recognition}, pages 2724--2732, 2018.

\bibitem[Panareda~Busto and Gall(2017)]{panareda2017open}
Pau Panareda~Busto and Juergen Gall.
\newblock Open set domain adaptation.
\newblock In \emph{Proceedings of the IEEE International Conference on Computer
  Vision}, pages 754--763, 2017.

\bibitem[Vinyals et~al.(2016)Vinyals, Blundell, Lillicrap, Wierstra,
  et~al.]{vinyals2016matching}
Oriol Vinyals, Charles Blundell, Timothy Lillicrap, Daan Wierstra, et~al.
\newblock Matching networks for one shot learning.
\newblock In \emph{Advances in neural information processing systems}, pages
  3630--3638, 2016.

\bibitem[Finn et~al.(2017)Finn, Abbeel, and Levine]{finn2017model}
Chelsea Finn, Pieter Abbeel, and Sergey Levine.
\newblock Model-agnostic meta-learning for fast adaptation of deep networks.
\newblock In \emph{Proceedings of the 34th International Conference on Machine
  Learning-Volume 70}, pages 1126--1135. JMLR. org, 2017.

\bibitem[Chen et~al.(2019)Chen, Liu, Kira, Wang, and Huang]{chen2018a}
Wei-Yu Chen, Yen-Cheng Liu, Zsolt Kira, Yu-Chiang~Frank Wang, and Jia-Bin
  Huang.
\newblock A closer look at few-shot classification.
\newblock In \emph{International Conference on Learning Representations}, 2019.

\bibitem[Scott et~al.(2018)Scott, Ridgeway, and Mozer]{scott2018adapted}
Tyler Scott, Karl Ridgeway, and Michael~C Mozer.
\newblock Adapted deep embeddings: A synthesis of methods for k-shot inductive
  transfer learning.
\newblock In \emph{Advances in Neural Information Processing Systems}, pages
  76--85, 2018.

\bibitem[Snell et~al.(2017)Snell, Swersky, and Zemel]{snell2017prototypical}
Jake Snell, Kevin Swersky, and Richard Zemel.
\newblock Prototypical networks for few-shot learning.
\newblock In \emph{Advances in Neural Information Processing Systems}, pages
  4077--4087, 2017.

\bibitem[Shao et~al.(2019)Shao, Lan, and Yuen]{shao2019regularized}
Rui Shao, Xiangyuan Lan, and Pong~C Yuen.
\newblock Regularized fine-grained meta face anti-spoofing.
\newblock \emph{arXiv preprint arXiv:1911.10771}, 2019.

\bibitem[Fang et~al.(2013)Fang, Xu, and Rockmore]{fang2013unbiased}
Chen Fang, Ye~Xu, and Daniel~N Rockmore.
\newblock Unbiased metric learning: On the utilization of multiple datasets and
  web images for softening bias.
\newblock In \emph{Proceedings of the IEEE International Conference on Computer
  Vision}, pages 1657--1664, 2013.

\bibitem[Li et~al.(2017)Li, Yang, Song, and Hospedales]{li2017deeper}
Da~Li, Yongxin Yang, Yi-Zhe Song, and Timothy~M Hospedales.
\newblock Deeper, broader and artier domain generalization.
\newblock In \emph{Proceedings of the IEEE international conference on computer
  vision}, pages 5542--5550, 2017.

\bibitem[Sharifi-Noghabi et~al.(2020)Sharifi-Noghabi, Peng, Zolotareva,
  Collins, and Ester]{hshn2020AITL}
Hossein Sharifi-Noghabi, Shuman Peng, Olga Zolotareva, Colin~C Collins, and
  Martin Ester.
\newblock Aitl: Adversarial inductive transfer learning with input and output
  space adaptation for pharmacogenomics.
\newblock \emph{Bioinformatics}, 36:\penalty0 i380--i388, 2020.

\end{thebibliography}
\end{document}